\documentclass[a4paper, 10pt, conference]{ieeeconf}      

\IEEEoverridecommandlockouts                              

\usepackage{amsmath} 
\usepackage{amssymb}  
\usepackage{cite}
\usepackage{graphicx}

\usepackage{soul,color} 
\usepackage[ruled,norelsize,linesnumbered,lined,commentsnumbered]{algorithm2e}

\makeatletter
\let\NAT@parse\undefined
\makeatother
\usepackage{hyperref}
\usepackage{cleveref}

\makeatletter
\newcommand{\removelatexerror}{\let\@latex@error\@gobble}
\makeatother

\newcommand\T{\rule{0pt}{2.4ex}}        
\newcommand\B{\rule[-1.2ex]{0pt}{0pt}}  
\title{\LARGE \bf
	Risky Action Recognition in Lane Change Video Clips using Deep Spatiotemporal Networks with Segmentation Mask Transfer 
}

\author{Ekim Yurtsever$^{*}$, Yongkang Liu$^{**}$, Jacob Lambert$^{*}$,  Chiyomi Miyajima$^{***}$,  Eijiro Takeuchi$^{*\dagger}$,\\ Kazuya Takeda$^{*\dagger}$ and John H. L. Hansen$^{**}$
	\thanks{$^{*}$E. Yurtsever, J. Lambert, E. Takeuchi and K. Takeda are with Nagoya University, Nagoya, Japan.
		}%
	\thanks{$^{**}$Y. Liu and J. H. L. Hansen are with UT Dallas, Texas, United States. 
	}%
	\thanks{$^{***}$C. Miyajima is with Daido University, Nagoya, Japan. 
	}%
	\thanks{$^{\dagger}$ E. Takeuchi and K. Takeda are also with Tier IV Inc., Nagoya, Japan.}
	\thanks{Corresponding author: Ekim Yurtsever, ekimyurtsever@gmail.com
	}%
}

\begin{document}

	\maketitle
	\thispagestyle{empty}
	\pagestyle{empty}

\begin{abstract}
	
	Advanced driver assistance and automated driving systems rely on risk estimation modules to predict and avoid dangerous situations. Current methods use expensive sensor setups and complex processing pipelines, limiting their availability and robustness. To address these issues, we introduce a novel deep learning based driving risk assessment framework for classifying dangerous lane change behavior in short video clips captured by a monocular camera. First, semantic segmentation masks were generated from individual video frames with a pre-trained Mask R-CNN model. Then, frames overlayed with these masks were fed into a time distributed CNN-LSTM network with a final softmax classification layer. This network was trained on a semi-naturalistic lane change dataset with annotated risk labels. A comprehensive comparison of state-of-the-art pre-trained feature extractors was carried out to find the best network layout and training strategy. The best result, with a 0.937 AUC score, was obtained with the proposed framework. Our code and trained models are available open-source\footnote{\label{footnote1}\href{https://github.com/Ekim-Yurtsever/DeepTL-Lane-Change-Classification}{https://github.com/Ekim-Yurtsever/DeepTL-Lane-Change-Classification}}.       
 
 
\end{abstract}

\section{Introduction}


Advanced Driver Assistance Systems (ADAS) and Automated Driving Systems (ADS) are being developed with the promise of reducing traffic and increasing safety on roads, translating to considerable economic benefits\cite{safe2018-av}. Automated driving functions categorized as level three and above have already seen some success, typically through lidar and radar perception, but the high cost of these sensing modalities has slowed their integration in consumer vehicles. Moreover, even though remarkable progress has been achieved, vehicles equipped with these technologies are still involved in traffic accidents \cite{tian2018deeptest}.

In contrast, camera-based solutions to challenging perception tasks are low-cost and increasingly robust. Developments in machine learning, particularly through deep convolutional neural networks (CNNs), significantly increased object detection capabilities \cite{Krizhevsky2012-qm} and made reliable object tracking achievable \cite{held2016learning}. Furthermore, CNNs trained on big datasets became capable of learning generic feature representations. As a result, generalized, multi-task networks were developed\cite{He2017-mr}, as well as end-to-end networks \cite{Bojarski2016-qq}, which avoid the need for complex pipelines. Combined with recurrent neural networks (RNNs) and specifically Long-Short Term Memory networks (LSTMs), spatiotemporal relationships can now be modeled for action recognition\cite{yue2015beyond}. In the vehicle domain, velocity estimation with neural networks using a monocular camera was achieved\cite{kampelmuhler2018camera}. As such, cameras, under their operational illumination conditions, become a feasible sensing modality for intelligent vehicle technologies. In this work, we propose a novel risk estimation system that uses vision as its sole modality. This increases the implementation possibility of our method, as cameras are inexpensive and readily available in everyday devices such as smartphones. 

\begin{figure}
	\centering
	\includegraphics[width=1\columnwidth]{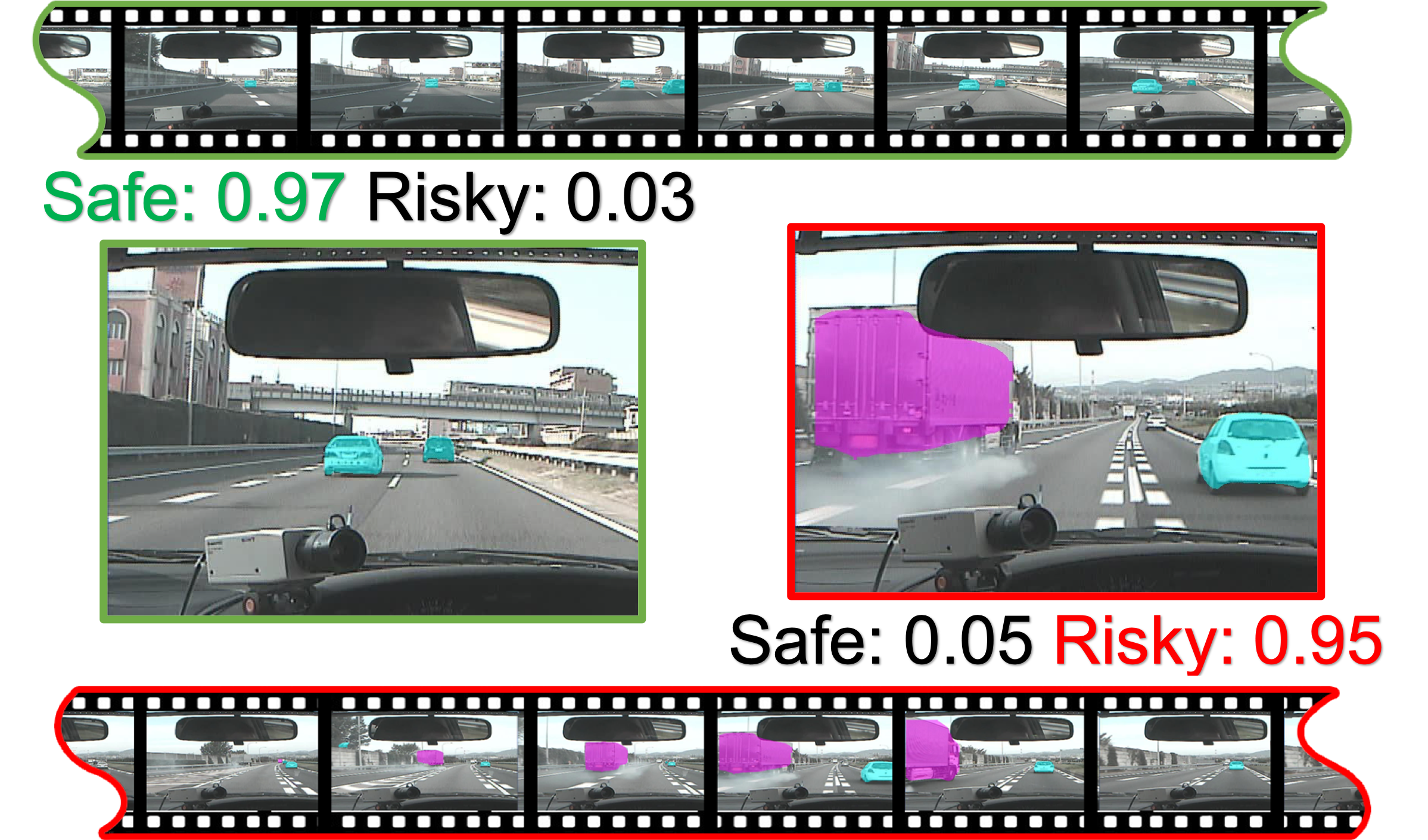}    
	\caption{Two lane change samples and classification outputs of our method. The average duration of these clips is $\sim$10 seconds. 0.937 AUC score was achieved with the proposed method. Video samples can be found in our repository\textsuperscript{\ref{footnote1}}.}
	\label{fig_safe_and_risky}
	\vspace{-15pt}
\end{figure}

The focus of this study is specifically risk estimation in lane changes. Lane changing is an essential driving action that is executed millions of times on a daily basis. It has been one of the most common pre-crash scenarios, where 7.62\% of all traffic accidents between light vehicles can be attributed to it \cite{najm2007pre}. Only rear-endings occur more frequently, which are primarily due to inattention. On the other hand, understanding a complex driving scene followed by acute decision making is necessary for negotiating a lane change. As such, a tool for unsafe behavior detection during lane change is of paramount importance. Heuristic and rule-based models often ignore the uncertainty of real systems. Forcing handcrafted policies or building deductive theories leads to observing unexpected behavior, which manifests itself as unmodeled dynamics in these approaches. As such, we believe a data-centric, learning based framework is imperative for finding the best explanation of the observed driving phenomena.


Our experiments evaluated several spatiotemporal network architectures on a naturalistic driving dataset consisting of 860 lane change videos. Individual sequences in the dataset were classified as risky or safe, as shown in \Cref{fig_safe_and_risky}. We also compared the feature representations of a wide selection of pre-trained state-of-the-art image classification networks. 

The major contributions of this work can be summarized as:
\begin{itemize}
	\item A novel deep learning based driving risk assessment framework with semantic mask transfer is proposed and used for detecting dangerous lane changes.
	\item Using solely a camera for the task 
	\item Extensive comparison of state-of-the-art deep backbone models with real-world data
\end{itemize}

The rest of the paper is organized as follows: after reviewing related literature in risk estimation, spatiotemporal classification and transfer learning, we describe the proposed method in \Cref{sec_method}. Then, the experimental setting is explained in \Cref{sec_experiments}, followed by results in \Cref{sec_results}.

\section{Related Works}
\subsection{Risk Studies}

Safety is a key factor driving intelligent vehicle technologies forward and an active area of research. Recently proposed ADAS usually attempt to detect and track surrounding objects and decide whether an action is necessary. Despite the successful implementation of such systems, there is no common agreement on the definition of risk. An objective definition based on a statistical probability of collision was proposed in \cite{Grayson2003-lh}. This \emph{objective risk} framework led to research focusing on vehicle tracking, where motion models predicted the future position of vehicles, allowing risk assessment\cite{Laugier2011-tx, Lefevre2014-cz}. On the other hand, risk metrics that consider the indeterministic human element was also proposed.  \emph{Subjective risk}, as in the risk perceived by human drivers, was studied as an alternative\cite{Fuller2005-tg}. The findings of this study indicate that human driving behavior is based on a perceived level of risk instead of calculated, objective collision probabilities.  Bottom-up unsupervised learning approaches were shown to be working for extracting individual driving styles\cite{yurtsever2019traffic}, but the latent learned representations were not associated with risk due to the nature of unsupervised learning.

Lane change is a typical driving maneuver that can be performed at a varied level of risk, and a significant percentage of all crashes happen due to the erroneous execution of it \cite{najm2007pre}. 
Risk in lane changes was studied mostly from the perspective of objective collision risk minimization\cite{Laugier2011-tx, woo2018adv}. A lane change dataset with manually annotated subjective risk labels made it possible to approach this problem from the perspective of supervised learning. The dataset includes ego-vehicle signals such as steering and pedal operation, range information and frames captured by a front-facing camera close to the drivers' point of view. However, previous works on this dataset ignored the monocular camera footage and used ego-vehicle signals\cite{yurtsever2018integrating, yamazaki2016integrating}. 

We utilized the video clips of the aforementioned dataset and focused solely on 2D vision in this study.


\subsection{Spatiotemporal Classification}

Image-based spatiotemporal classification research primarily focuses on video classification. An active application in this field, closely related to this work, is action recognition, where a sequence of 2D frames must be classified into one of many, some very similar, actions. The widely used UCF101 action recognition dataset\cite{soomro2012-np} features 101 actions such as running or soccer penalty kicks, which are difficult to differentiate when examining individual frames. Another widely used dataset is the Sports1M dataset\cite{KarpathyCVPR14}, which features one million videos of 487 classes of sports-related actions. 

The spatial relationship of things forms the context of a single image frame. The spatiotemporal context, on the other hand, is constituted by the motion of things, spanned across time in multiple frames. This makes action recognition a more challenging problem than image classification. Furthermore, in the case of a moving data collection platform, such as a vehicle equipped with a camera, distinguishing the local spatiotemporal context (the motion of things \textit{in} the scene), apart from the global context (the motion \textit{of} the scene), increases the difficulty of the problem.


While video classification has a history of using traditional computer vision, the current state-of-the-art is entirely dominated by deep learning approaches. Transfer learning is a staple in these methods: CNNs pre-trained on large image datasets are used as a starting point, then modified for spatiotemporal classification and fine-tuned using action recognition datasets. Early work introduced two network archetypes, one in which spatial and temporal features were extracted simultaneously by a single network\cite{KarpathyCVPR14} and the other which had two distinct spatial and temporal branches followed by fusion\cite{Simonyan2014-cb}. Recently, both 3D CNNs\cite{Diba2017-dn} as well as RNN variants, especially LSTMs, have been used to approach this problem\cite{Donahue2017-vb}. Various 2D CNNs have been shown to produce good input features for temporal networks like LSTMs. Optical flow output from CNNs has also been used as input to LSTMs\cite{yue2015beyond}. Closely related to this work, features extracted from deep CNNs have been used as input to temporal networks \cite{Chao2018-qr, Girdhar2018-tc}.

\subsection{Transfer Learning}
Transfer learning can be generally thought of as modifying an existing network for some other application. More precisely, given a source domain and learning task, the aim is to transfer the source model's knowledge to a target model, which may have another target domain and task. Transfer learning methods are further classified depending on how the source and target domain and task differ. Inductive transfer learning refers to the case where the source and target tasks are different, whereas in transductive transfer learning, the domain changes while the task remains the same\cite{Pan2010-rp}.

Transfer learning has been used for diverse applications with varying complexity. It has been used for natural language processing\cite{Howard2018-un}, speech recognition across different languages\cite{Huang2013-wx}, voice conversion\cite{tobing2019voice} and other computer vision applications\cite{Sunderhauf2015-at}. The instance segmentation algorithm Mask R-CNN\cite{He2017-mr}, which is used in this paper, demonstrated the benefits of multi-task transfer learning. It was trained for both bounding box estimation and instance segmentation, yet it was shown to outperform its previous work which focuses on the former\cite{Ren2015-bq}. Furthermore, when its knowledge was transferred to the task domain of human pose estimation, it significantly outperformed competing algorithms. This shows the potential of transfer learning for network generalization, which was leveraged in this research. 

Inductive transfer learning was used in this work, as the target domain is somewhat similar to the source domain, but the target task is entirely different. Specifically, the assumption tested here is that feature representations learned by deep CNNs trained on large image databases should be transferable to our task and domain: classifying lane change video clips as safe or dangerous. Feature representations obtained from several pre-trained networks were utilized to test this assumption as outlined in \Cref{sec_transfer}.

\section{Proposed Method}\label{sec_method}

A novel deep spatiotemporal driving risk assessment framework with Semantic Mask Transfer (SMT) is proposed here. The proposed strategy was used for recognizing risky actions in short lane change clips. Furthermore, an exhaustive comparison of state-of-the-art deep feature extractors was carried out to find the best model layout.


\subsection{Problem Formulation}

The principal postulation of this study is the partition of the whole lane change set into two jointly exhaustive and mutually exclusive subsets: safe and risky. This proposition is an oversimplification and, depending on the domain, may not suffice. Nevertheless, this dichotomy simplifies problem formulation and enables the employment of state-of-the-art binary video classification methods. All hypotheses that contradict with this postulation are out of this study's scope.

The objective is to classify a sequence of images captured during a lane change into the risky or the safe subset. The temporal dimension of videos can vary depending on the application and the lane change itself. However, a fixed number of frame constitution is assumed in this study. This decision enables the deployment of fixed-dimension network architectures to solve the problem. The classification problem is formulated as follows:

For lane change $i$, the goal is to find the inferred risk label $\hat{y}_{i}$, given the sequence of images captured during the lane change $\textbf{x}_{i} = (x_{1},x_{2}, \cdots, x_{T}) $, with the spatiotemporal classification function $f$. 
\begin{equation}
\label{eqn_classification}
\hat{y_{i}} = f(\textbf{x}_{i})
\end{equation}

%

\begin{figure*}%
	\centering
	\includegraphics[width=1.6\columnwidth]{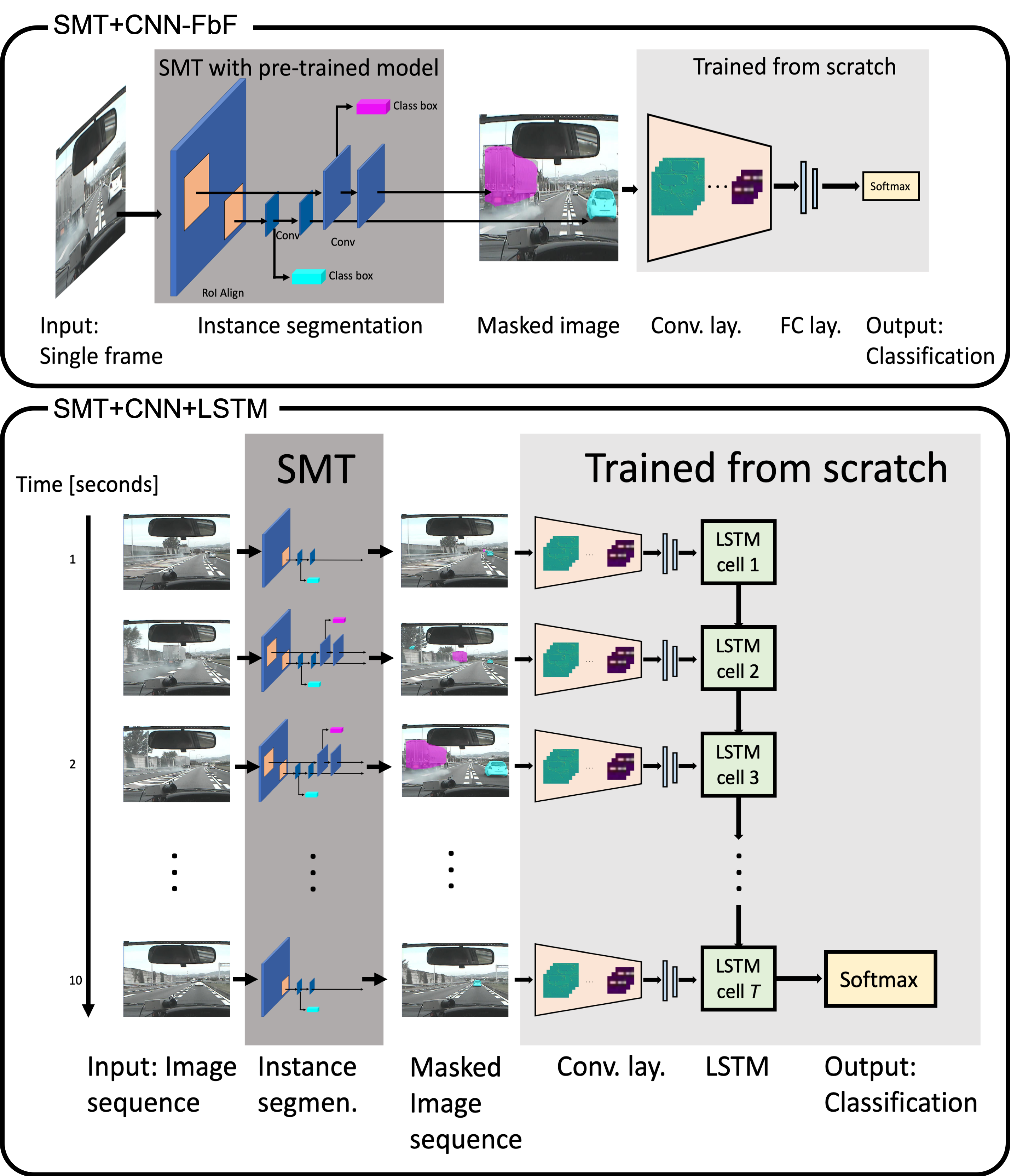}
	
	\caption{The proposed framework. SMT stands for semantic mask transfer. Mask R-CNN\cite{He2017-mr} is used as the mask extractor, and no fine-tuning is done for the SMT part. In this implementation, trucks are colored in magenta and cars in cyan. The average duration of lane change clips is $\sim 10$ seconds. Each clip is subsampled before processing. Details of frame selection is shown in Figure \ref{fig_uniform_frame}. A temporal composition with contrasting elements can be more useful for relaying semantics of the scene. In our subjective opinion, after a glance, the masked image sequence relays a more striking version of the lane change action than the raw sequence.  }
	\label{fig_architecture2}
	\vspace{-15pt}
\end{figure*}
\noindent where $T$ is fixed $\forall$ lane changes and risk label $y$ is encoded as a one-hot vector.
\begin{equation}
\label{eqn_binary_labels}
y=
\begin{cases} 
(1, 0) & \text{for safe lane changes}\\
(0, 1) & \text{for risky lane changes}
\end{cases}.
\end{equation}

The spatiotemporal classifier, $f$, is learned with supervised deep learning models. Extraction of the ground truth, $y$, is explained in Section \ref{sec_ground_truth}

\subsection{Deep Neural Network Architectures}

Besides the proposed framework, spatiotemporal classification with semantic mask transfer (SMT+CNN+LSTM), a significant contribution of this study is the comprehensive experimental analysis and evaluation of the state-of-the-art video classification architectures for the task at hand. 

Two different learning strategies were followed in this work. The first one was the conventional supervised deep learning approach: training a deep neural network architecture from scratch with raw image input and target risk labels through backpropagation. In the second approach, transfer learning was utilized for extracting high-level abstract features from the raw image data. After extraction, these features were fed into separate classifiers which were trained using the target risk labels. A wide selection of pre-trained state-of-the-art very deep networks was used as feature extractors in the experiments.

Furthermore, six architecture families were compared throughout the experiments. Details of each are given in the following sections. High-level diagram of the proposed method, labeled as SMT+CNN+LSTM, is shown in Figure \ref{fig_architecture2}.  


\subsection{Training From Scratch}\label{training_scratch}

Deep learning is a popular machine-learning algorithm family. It is widely used especially for computer vision tasks. However, huge amounts of data are required to train deep architectures. Without adequate data, the performance drops significantly. The lane change dichotomy that is introduced here is not a well-established domain in comparison to the standard image classification problem. As such, big datasets that are annotated laboriously such as ImageNet\cite{deng2009imagenet} and COCO\cite{lin2014microsoft} do not exist for the task. Therefore, a specific lane change dataset was collected and annotated\cite{takeda2011international}. The scope of this corpus, however, is not on the same scale as the mentioned datasets. 

Two different architectures were used for the training from scratch stratagem. CNNs are an integral part of state-of-the-art image classification models. As such, Frame-by-Frame (FbF) classification with CNNs was selected as the baseline here. The baseline was compared against the state-of-the-art spatiotemporal classification architecture; the CNN + Long Short-Term Memory (CNN+LSTM) model.  

\textbf{Frame-by-frame classification with CNNs (FbF CNN)}

A CNN architecture with a fully connected softmax final layer was designed as the baseline in this study. The temporal dimension of lane change clips was disregarded in the baseline. In other words, each frame was classified independently as safe or risky. 

The architecture is given in shorthand notation as follows: $x_{j} \rightarrow C(64, 5, 1)\rightarrow P\rightarrow C(32, 5, 1)\rightarrow P\rightarrow \text{FC}(1000)\rightarrow \text{Softmax}(2)\rightarrow\hat{y}_{j}$. Where $C(r, w, s)$ indicates a convolutional layer with $r$ filters, a $w\times w$ window and $s$ stride size. $P$ stands for max pooling layers and $\text{FC}(h)$ for a fully connected dense layer with $h$ hidden units. The final layer is a fully connected softmax with 2 classes. In order to train this network, risk labels were replicated to each constituent frame correspondingly:     
\begin{equation}
\label{eqn_CNN-FbF}
\forall x_{j} \in \textbf{x}, \text{ } y_{j} = y
\end{equation}

\noindent where $x_{j}$ is the $j$th frame of the lane change $i$.

\textbf{Spatiotemporal classification (CNN+LSTM)}

Video classification is a spatiotemporal problem. Therefore, architectures that consider spatial and temporal aspects of the input data are expected to perform better than the baseline.  

State-of-the-art performance for action clip classification was achieved in \cite{yue2015beyond} with a Long-Short Term Memory (LSTM) network where CNN features were fed as inputs of each time step. The distinction between two separate actions such as walking and jumping \textit{might} be easier to detect than a postulated difference such as a risky or safe lane change. However, even though action clip classification is not an identical problem to the lane change dichotomy, it is still relevant and shares the same modality.  As such, a similar architecture is proposed here to solve the problem at hand as follows:
\begin{equation}
\label{eqn_CNN+LSTM}
\hat{y} = f_{\text{LSTM}}(f_{\text{CNN}}(x_{1}), \cdots, f_{\text{CNN}}(x_{j}), \cdots,f_{\text{CNN}}(x_{T})).
\end{equation}

The spatial feature sequence $\textbf{z} = (z_{1}, z_{2}, \cdots, z_{T})$ of a given lane change was extracted from the raw image sequence $\textbf{x}$ with the CNN feature extractor $f_{\text{CNN}}$. $\textbf{z}$ was then fed into LSTM cells to infer $\hat{y}$. 

The LSTM network computes the sequence of hidden vectors $\textbf{h} = (h_{1}, h_{2}, ..., h_{T})$ given the input feature sequence  $\textbf{z}$ by iterating the following equations for each timestep $t$.

\begin{equation}
\label{eqn_LSTM_2}
g_{t} = \sigma(W_{g}z_{i, t} + U_{g}h_{t-1}+b_{g}),
\end{equation}
\begin{equation}
\label{eqn_LSTM_3}
i_{t} = \sigma(W_{i}z_{i, t} + U_{i}h_{t-1}+b_{i}),
\end{equation}
\begin{equation}
\label{eqn_LSTM_4}
o_{t} = \sigma(W_{o}z_{i, t} + U_{o}h_{t-1}+b_{o}),
\end{equation}
\begin{equation}
\label{eqn_LSTM_5}
c_{t} = g_{t} \circ c_{t-1} + i_{t} \circ \text{tanh}(W_{c}z_{i, t}+ U_{c}h_{t-1}+b_{c}),
\end{equation}
\begin{equation}
\label{eqn_LSTM_6}
h_{t} = o_{t} \circ \text{tanh}(c_{t}),
\end{equation}

\noindent where $g_{t}$, $i_{t}$, $o_{t}$ are the activation functions of the forget gate, the input gate and the output gate respectively. $c_{t}$ is the cell state vector, $\circ$ is Hadamard product i.e element-wise product and $W$, $U$, $b$ are weight matrices that are learned through training. $\sigma$ is the sigmoid activation function.

A many-to-one layout was used as only one label is required per lane change. The last output vector of the LSTM, $h_{T}$, was fed into a dense layer with a softmax activation function to infer the risk label $\hat{y} = f_{\text{softmax}}(h_{T})$. The shorthand notation of the complete architecture is as follows: $\textbf{x}\rightarrow x_{j}\rightarrow C(16, 3, 1)\rightarrow C(16, 3, 1)\rightarrow P\rightarrow D\rightarrow \text{FC}(200)\rightarrow\text{FC}(50)\rightarrow z_{j}\rightarrow\textbf{z}\rightarrow\text{LSTM}(q, 20)\rightarrow\text{Softmax}(2)\rightarrow\hat{y}$. $D$ stands for a dropout layer with 0.2 dropout probability. LSTM$(q, h)$ indicates an LSTM layer with $q$ time steps and $h$ hidden units. $q$ was changed throughout the experiments. Details of the temporal dimension is given in Section \ref{sec_temp_dim}.

\subsection{Transfer Learning}\label{sec_transfer}
\begin{figure*}%
	\centering
	\includegraphics[width=2\columnwidth]{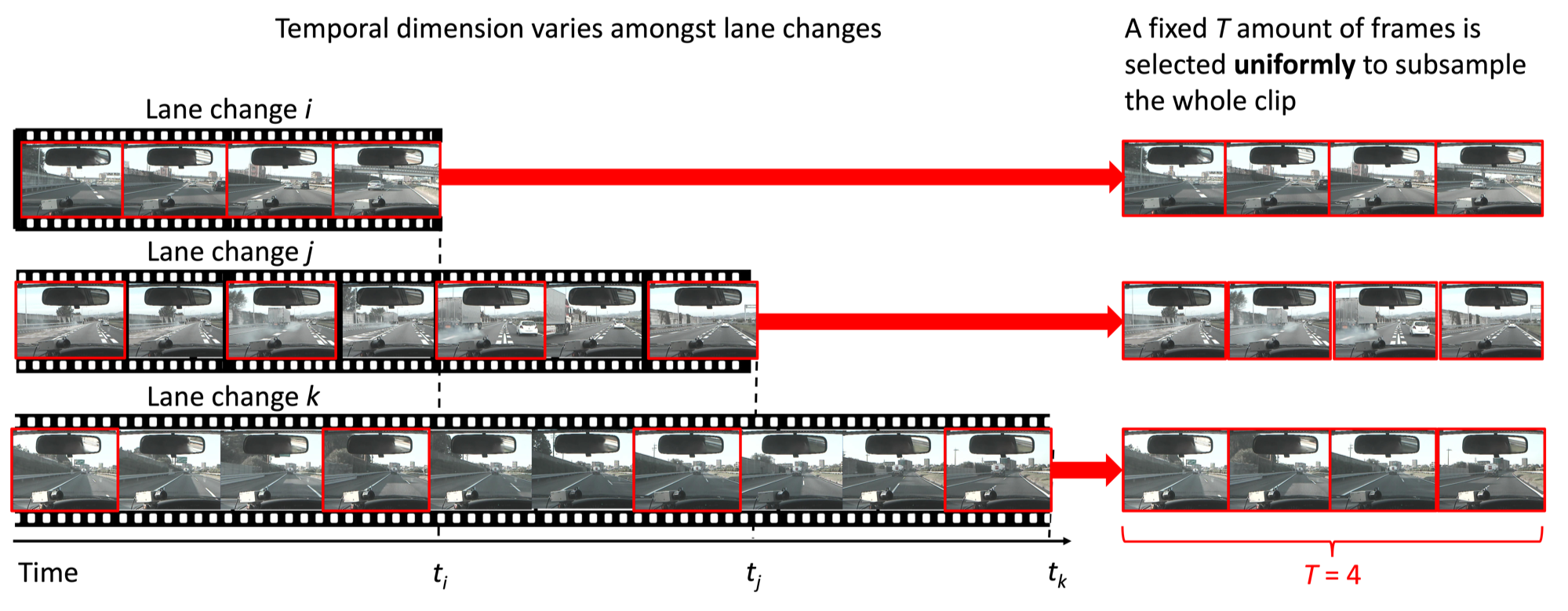}
	
	\caption{Subsampling the video clips. A fixed amount of $T$ frames is selected uniformly for each lane change. This design choice enables the employment of fixed frame rate architectures. $T$ is a hyper-parameter of our framework and it affects the classification performance. We tested different values of $T$ throughout the experiments. This point is further elaborated in Section \ref{sec_results}.  }
	\label{fig_uniform_frame}
	\vspace{-15pt}
\end{figure*}

As mentioned earlier, supervised training of very deep networks requires enormous amounts of data. This creates a bottleneck for certain problems such as the lane change dichotomy due to the lack of a big dataset. This issue can be circumvented with the use of models that are pre-trained on big datasets. Even though the target task, classifying a sequence of lane change images as risky or safe, is different from the source task of classifying a single image as one of the thousands of classes of ImageNet\cite{deng2009imagenet} dataset, pre-trained state-of-the-art networks can be utilized as feature extractors. In this study, four different transfer learning architectures were compared.


\textbf{Frame-by-frame classification with feature transfer (FbF FT)}

Frame-by-frame classification with feature extraction is accepted as the baseline transfer learning strategy of this study. The method is straightforward: first, the pre-trained very deep network was cut before its final fully connected layer. Then, for each frame $x_{j}$, the transferred spatial feature $z_{j}$ was obtained with the truncated pre-trained network $f_{t}$.

\begin{equation}
\label{eqn_transfer_FbF}
z_{j} = f_{t}(x_{j}).
\end{equation}

Finally, the extracted feature $z_{j}$ was fed into a shallow fully connected softmax classifier that was trained with the lane change data to infer the risk labels.  VGG19\cite{Simonyan2015-nf}, MobileNet\cite{Howard2017-rr}, InceptionResNet\cite{Szegedy2017-yz}, NasNET\cite{Zoph2018-yj}, Xception\cite{Chollet2017-sq} and ResNet\cite{He2016-db} were used as feature extractors in the experiments. All of the networks were pre-trained on the ImageNet\cite{deng2009imagenet} dataset.

\textbf{Spatiotemporal classification with feature transfer (FT+LSTM)}

The second strategy had the same spatial feature extraction step, but a full temporal network was trained instead of a shallow classifier with the lane change data. The same pre-trained networks used for the FbF FT were utilized again for feature extraction.

In summary, FbF FT and FT+LSTM are similar to the training from scratch strategies, namely FbF CNN and CNN+LSTM. The only difference is the replacement of training of convolutional layers with feature transfer. 

\textbf{Frame-by-frame classification with semantic mask transfer (FbF SMT+CNN)}

Multi-task deep networks have become popular recently, especially for vision tasks, in urban driving applications.  State-of-the-art multi-task networks YOLOv3\cite{Redmon2018-ar} and Mask R-CNN\cite{He2017-mr} were used for segmentation mask transfer in this study. Both of the networks were pre-trained on the COCO\cite{lin2014microsoft} dataset. 

The performance of the pre-trained Mask R-CNN can be qualitatively analyzed by inspecting Figure \ref{fig_architecture2}. The segmentation masks shown in the figure were obtained for the lane change dataset \textit{without} any fine-tuning or training. It is the out-of-the-box performance of Mask R-CNN trained on the COCO dataset, with our inputs. 

A slight post-process modification was done to YOLOv3 in order to obtain segmentation masks. The original network outputs a bounding box and class id for detected objects. For each class, bounding boxes were filled with 0.7 opacity and with a unique color in this study. Mask R-CNN did not need any modification as it outputs a segmentation mask besides bounding box and class id. It is assumed that a high-contrast composition is more useful for discerning distinct elements. As such, bright and unnatural colors such as cyan and magenta were selected as mask colors.

The out-of-the-box performance was good but not perfect. Since the ground truth for segmentation masks are not available for the lane change dataset, fine-tuning was not possible nor any quantitative analysis. However, these masks can still be used for risk detection. The idea is very similar to FbF CNN: first, the lane change frames were converted to the masked images with pre-trained networks. After this step, the same network architecture of FbF CNN was used to infer risk labels.   

\textbf{Spatiotemporal classification with semantic mask transfer (SMT+CNN+LSTM)}

The main contribution of this work, SMT+CNN+LSTM, is a novel framework for binary video classification. To the best of authors' knowledge, semantic segmentation masks had not been fed into an LSTM architecture for risk detection in the literature before. The proposed method is shown in Figure \ref{fig_architecture2}.

The main hypothesis is that a temporal composition with highly contrasting elements can tell a better story. Qualitative evaluation of this claim can be done by inspecting Figure \ref{fig_architecture2}. In our subjective opinion, after a glance, the masked image sequence relays a more striking version of the lane change action than the raw sequence. Quantitative analysis is given in Section \ref{sec_results}.

The starting point of the framework is the masked images whose extraction is described in the previous section. These masked images were passed through convolutional layers to extract abstract features from the \textit{contrasted compositions} created by the mask colors. These high-level features were then fed into LSTM cells to depict temporal relationships. The same CNN+LSTM architecture that is given in Section \ref{training_scratch} was used with the only difference being the input type. SMT+CNN+LSTM uses images with overlayed semantic segmentation masks as its sequential input.


\section{Experiments}\label{sec_experiments}

\subsection{Dataset}\label{sec_data_collection}\label{sec_ground_truth}

%
A subset of the NUDrive dataset was used in this study. Data was collected with an instrumented test vehicle in Nagoya, Japan. Details of the corpus can be found in \cite{takeda2011international}.
The subset consists of 860 lane change video clips captured by a front-facing camera. Eleven different drivers executed the lane changes on Nagoya expressway. Drivers followed the same route and were asked to keep their natural driving habits while doing lane changes as much as possible. The whole trip of each driver was parsed manually to extract the lane change clips afterward. 

The footage was captured with a resolution of 692x480 at 29.4 frames per second. The average duration of a lane change clip is approximately 10 seconds. 

\textbf{Establishing the ground truth:} Ten annotators watched the video clips and rated the risk level of each instance subjectively. Annotators gave a risk score between one (safest) and five (most risky) to each lane change. Risk ratings were normalized for each annotator. Then, the normalized scores obtained from ten annotators were averaged to obtain a single score per lane change. The riskiest 5\% of the lane change population was accepted as risky while the rest was assumed to be safe. Risky lane changes were taken as the positive class in this binary classification. The final distribution is 43 to 817 for the positive and negative classes respectively.

\begin{table*}[t]
	\caption{Classification performances on the lane change dataset. Our method, smt+cnn+lstm, achieved the best AUC score.  }
	\vspace{-3.6mm}
	\label{table_risk_auc}
	\begin{center}
		\begin{tabular*}{\textwidth}{c @{\extracolsep{\fill}} cccccc}
			\hline\hline \T
			Architecture & Backbone model & Pre-trained on & $ \# $ Parameters (millions) & Best $T$ &AUC \\
			\hline\T 
			FbF CNN          & -               & -         & 1.6 & 1  & 0.815 \\
			FbF FT           & VGG19 \cite{Simonyan2015-nf}           & ImageNET\cite{deng2009imagenet}  & 144.1 & 1  & 0.809 \\
			FbF FT           & MobileNet \cite{Howard2017-rr}       & ImageNET   &3.7 & 1  & 0.779 \\
			FbF FT           & Inceptionresnet \cite{Szegedy2017-yz} & ImageNET   & 56 & 1  & 0.617 \\
			FbF FT           & NASNet \cite{Zoph2018-yj} & ImageNET   &89.5     & 1  & 0.738 \\
			FbF FT           & Xception \cite{Chollet2017-sq}       & ImageNET  &23.1 & 1  & 0.683 \\
			FbF FT           & ResNet50 \cite{He2016-db}        & ImageNET   & 25.8 & 1  & 0.861 \\
			FbF SMT+CNN     & YOLOv3 \cite{Redmon2018-ar}  & COCO\cite{lin2014microsoft}       &63.6 & 1  & 0.854 \\
			FbF SMT+CNN    & Mask RCNN \cite{He2017-mr}       & COCO       & 65.8 & 1  & 0.853 \\
			CNN+LSTM       & -               & -          & 0.6 & 50 & 0.888 \\
			FT+LSTM        & VGG19 \cite{Simonyan2015-nf}           & ImageNET   & 143.9 & 50 & 0.886 \\
			FT+LSTM        & MobileNet \cite{Howard2017-rr}       & ImageNET   & 3.6 & 10 & 0.844 \\
			FT+LSTM        & Inceptionresnet \cite{Szegedy2017-yz}      & ImageNET   & 56 & 20 & 0.5   \\
			FT+LSTM        & NASNet \cite{Zoph2018-yj} & ImageNET   & 89.3 & 5 & 0.761 \\
			FT+LSTM        & Xception \cite{Chollet2017-sq}       & ImageNET   & 23 & 50 & 0.768 \B \\
			\hspace{-21mm}\textbf{Best 3} \\
			\hline \T 
			FT+LSTM& ResNet50 \cite{He2016-db}        & ImageNET  & 25.8 & 20 & 0.910 \\
			SMT+CNN+LSTM & YOLOv3 \cite{Redmon2018-ar} & COCO      & 62.5 & 50 & 0.927 \\
			\textbf{SMT+CNN+LSTM} & \textbf{Mask R-CNN}\cite{He2017-mr}       & COCO      & 64.8 & 50 & \textbf{0.937} \\
			\hline\hline
			
		\end{tabular*}
		
	\end{center}    
	FbF: Frame-by-frame, FT: Feature Transfer, SMT: Semantic Mask Transfer
\end{table*}

\subsection{Experimental Conditions and Evaluation Criteria}\label{sec_temp_dim}\label{sec_evaluation}

Temporal dimension length, which is equal to the number of frames fed into the LSTM architecture, was changed throughout the experiments. The number of frames affected the performance significantly. Details of this phenomena are discussed further in Section \ref{sec_results}. 

For each architecture and cross-validation fold, seven different training sessions were run with 5, 10, 15, 20, 50, and 100 frames that were subsampled uniformly per lane change sequence. The average total number of frames per lane change is around 300. The uniform video subsampling is shown in Figure \ref{fig_uniform_frame}. 

10-fold-cross-validation was applied all through the experiments. 18 architectures and 6 subsampling options were compared with 10-fold-validation, which totaled in training of 1080 networks.

The lane change dataset is heavily skewed towards the negative class. Accuracy is not a definitive metric under this circumstance. Instead, Area Under the Curve (AUC) was chosen as it is widely used for binary classification problems with large class imbalance. The evaluation focus of AUC is the ability for avoiding false classification \cite{sokolova2009systematic}. Besides classification performance, inference time is an important criterion. Especially for real-time applications very deep networks can get cumbersome. The main factor that affects inference time is the number of total parameters in an architecture. The evaluation of the experiments with respect to these criteria is given in Section \ref{sec_results}.

\subsection{Training and Implementation Details}

The Adam optimizer was used throughout the experiments with 0.0001 learning and 0.01 decay rate. A batch size of 32 was used on each training run which consisted of 1000 epochs. Training to validation split-ratio was 0.9 for all cross-validation runs. A categorical cross entropy loss function was employed for all architectures. 

The proposed approaches were implemented in Keras, a deep learning library for Python. Our code is open-source and can be accessed from our GitHub repository\footnote{\href{https://github.com/Ekim-Yurtsever/DeepTL-Lane-Change-Classification}{https://github.com/Ekim-Yurtsever/DeepTL-Lane-Change-Classification}}.
The computational experiments took less than a month to finish. A GPU cluster with 6 Nvidia GTX TITAN X was utilized for this research. 

\section{Results}\label{sec_results}

Table \ref{table_risk_auc} summarizes the experimental results, where network architectures are shown in the first column. The backbone model column indicates the base transferred very deep network if there was any. Not all architectures used transfer learning, namely FbF CNN and CNN+LSTM. Datasets that the transferred networks were pre-trained on are given in column three. It should be noted again that all architectures were trained with our data for the final classification task. The total number of network parameters for each architecture is shown in column four for assessing the computational load. Lower parameter amount correlates with faster inference time. $T$, the fixed number of frames, were changed between 5 to 100 for each configuration. The best scoring $T$ in terms of AUC of each row  is given in column five.  The final column is the AUC score, the main performance metric of this study.  

All spatiotemporal architectures with an LSTM layer outperformed their spatial counterparts, except the configuration with the Inceptionresnet\cite{Szegedy2017-yz} base model, which had the lowest performance. These results underline the importance of the temporal dimension. However, a very large dependence on temporal information is also undesired because; it swells the network, increases the input data size and slows the inference time.       

The best result was obtained with the proposed SMT+CNN+LSTM framework which used a Mask R-CNN\cite{He2017-mr} semantic mask extractor. We believe this result was due to the masked-contrasted temporal compositions' aptitude for relaying semantic information. The third best result was obtained with an FT+LSTM architecture which used ResNet50\cite{He2016-db} as its backbone model. The rest of the architectures fell behind the top three by a noticeable margin. For example, the proposed SMT+CNN+LSTM's risky lane change detection performance was 25\% better than the FbF FT with an Xception backbone. 

\section{Conclusions}


Classifying short lane change video clips as risky or safe has been achieved with a 0.937 AUC score using the proposed SMT+CNN+LSTM method. 

Our experiments bolster the belief in the adaptive capabilities of deep learning. Transfer learning expands the potential use of trained models. With the increasing availability of open-source libraries and fully trained models with high out-of-the-box performance, new problems can be tackled without tailoring huge datasets for them. The results of this study reinforce this claim.


Promising results were obtained in this work, but only a single driving action, the lane change maneuver, was investigated. In order to parse and assess continuous driving footage, more spatiotemporal techniques should be tested such as feature pooling and 3D convolution in future works. Improving the transfer learning strategies with fine-tuning and utilizing more modalities such as lidar are also amongst our future objectives.






\section*{ACKNOWLEDGMENT}
    
This work has been partly supported by the New Energy and Industrial Technology Development Organization (NEDO).

\bibliographystyle{IEEEtran}
\bibliography{references_lane_change_risk}

\end{document}